\documentclass[10pt,twocolumn,letterpaper]{article}

\usepackage[pagenumbers]{cvpr} 

\usepackage{graphicx}
\usepackage{amsmath}
\usepackage{amssymb}
\usepackage{booktabs}

\usepackage{multirow}
\usepackage{xcolor}
\usepackage{array}
\usepackage{soul}
\usepackage{enumitem}

\usepackage{float}
\usepackage{cite}

\usepackage[pagebackref,breaklinks,colorlinks]{hyperref}

\usepackage[capitalize]{cleveref}

\newcolumntype{P}[1]{>{\centering\arraybackslash}p{#1}}
\newcolumntype{M}[1]{>{\centering\arraybackslash}m{#1}}
\newcolumntype{H}{>{\setbox0=\hbox\bgroup}c<{\egroup}@{}}

\crefname{section}{Sec.}{Secs.}
\Crefname{section}{Section}{Sections}
\Crefname{table}{Table}{Tables}
\crefname{table}{Tab.}{Tabs.}

\newcommand{\netone}{{\emph{Pose2PersonNet}}\@\xspace}
\newcommand{\nettwo}{{\emph{Person2SceneNet}}\@\xspace}
\newcommand{\stageone}{{\emph{Pose2Person}}\@\xspace}
\newcommand{\stagetwo}{{\emph{Person2Scene}}\@\xspace}

\newcommand{\methodname}{IncogniMOT}

\begin{document}

\title{
Data-Driven but Privacy-Conscious:\\Pedestrian Dataset De-identification via Full-Body Person Synthesis
}

\author{
Maxim Maximov{\textbf{\textsuperscript{*}}}\\
TU Munich\\
\and
Tim Meinhardt{\textbf{\textsuperscript{*}}}\\
TU Munich\\
\and
Ismail Elezi\\
TU Munich\\
\and
Zoe Papakipos\\
Meta AI\\
\and
Caner Hazirbas\\
Meta AI\\
\and
Cristian Canton\\
Meta AI\\
\and
Laura Leal-Taixé\\
TU Munich{\textbf{\textsuperscript{**}}}\\
}

\maketitle

\begin{abstract}
    The advent of data-driven technology solutions is accompanied by an increasing concern with data privacy.
    This is of particular importance for human-centered image recognition tasks, such as pedestrian detection, re-identification, and tracking.
    To highlight the importance of privacy issues and motivate future research, we motivate and introduce the Pedestrian Dataset De-Identification (PDI) task.
    PDI evaluates the degree of de-identification and downstream task training performance for a given de-identification method.
    As a first baseline, we propose \methodname, a two-stage full-body de-identification pipeline based on image synthesis via generative adversarial networks.
    The first stage replaces target pedestrians with synthetic identities.
    To improve downstream task performance, we then apply stage two which blends and adapts the synthetic image parts into the data.
    To demonstrate the effectiveness of \methodname, we generate a fully de-identified version of the MOT17 pedestrian tracking dataset and analyze its application as training data for pedestrian re-identification, detection, and tracking models. 
    Furthermore, we show how our data is able to narrow the synthetic-to-real performance gap in a privacy-conscious manner.
\end{abstract}

\makeatletter{\renewcommand*{\@makefnmark}{}
\footnotetext{* Authors contributed equally.}\makeatother}

\makeatletter{\renewcommand*{\@makefnmark}{}
\footnotetext{** Currently at NVIDIA.}\makeatother}

\section{Introduction}
\label{sec:intro}

\begin{figure}[t]
    \centering
    \includegraphics[width=0.8\columnwidth, trim={10px 175px 370px 0px}, clip]
    {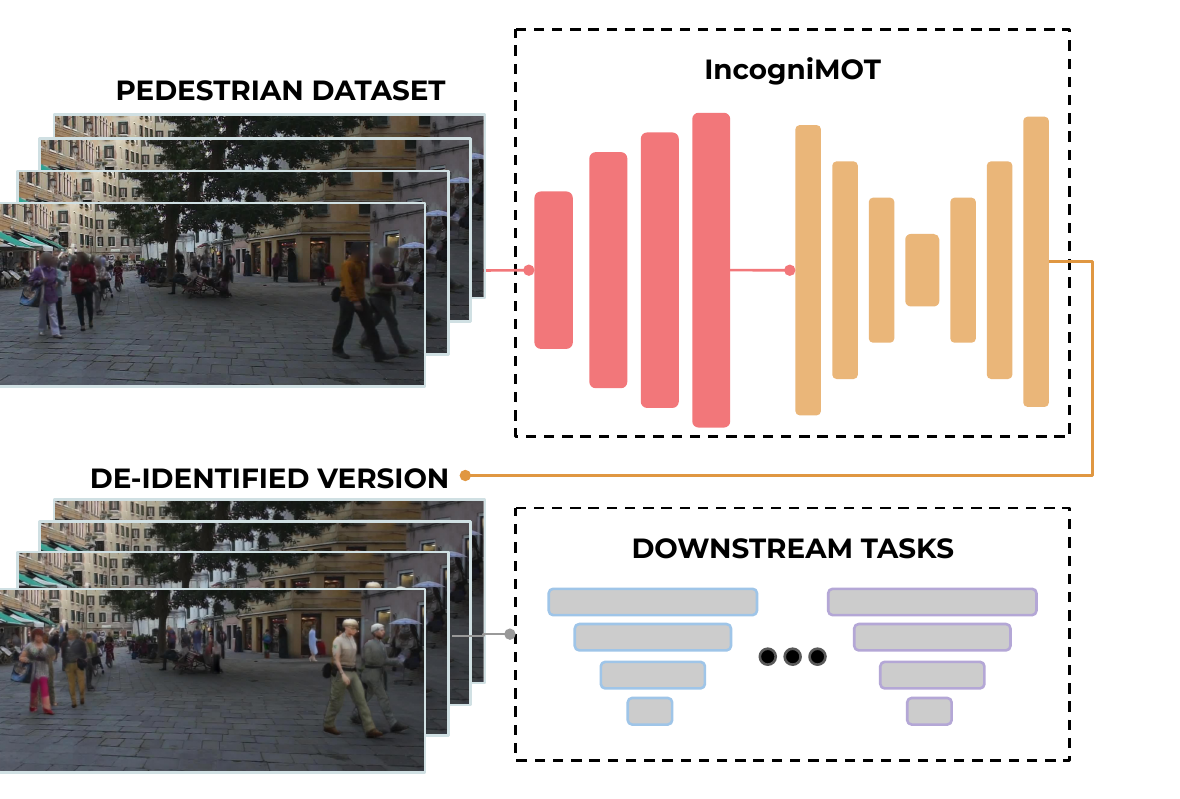}
    
    \caption{
        To address the increasing concerns with data consent and privacy, we present the Pedestrian Dataset De-Identification task and \methodname, a full-body two-stage de-identification pipeline.
        }
        
        \vspace{-.5cm}
    \label{fig:teaser}
\end{figure}

Over the last decade, the computer vision research community has achieved impressive breakthroughs on a range of challenging image and video recognition tasks.
These breakthroughs were largely driven by the increased availability of computational resources and large-scale datasets, both of which allow modern machine learning algorithms to unfold their potential.
Nowadays, state-of-the-art deep learning models even surpass human-level performance~\cite{DBLP:journals/corr/HeZR015} on complex tasks such as image classification or object detection.
As these machine recognition systems keep improving, new potential for applications in emerging fields is unlocked.
These applications generally revolve around automation, for example in robotics, autonomous driving, augmented/virtual reality, or health/sports analysis.

Large-scale dataset training is at the center of computer vision research and a key component of most modern vision pipelines.
However, for recognition tasks centered around humans, e.g., person detection, tracking, pose estimation, data collection is accompanied by largely unaddressed concerns regarding \emph{data privacy}.
In particular, the collection of images and videos of people without their explicit consent raises serious ethical questions.
The \emph{General Data Protection Regulations} (GDPR)~\cite{EUdataregulations2018}  approved by the European Union in 2018 put these concerns into law by prohibiting the release of any unlawfully acquired data.
In computer vision, two person/face recognition datasets, DukeMTMC~\cite{ristani2016MTMC} and MS-Celeb-1M \cite{DBLP:conf/eccv/GuoZHHG16}, had to be retracted after it was published that their data was used to train models for several controversial applications.

Collecting human-centered datasets, especially with crowded and public scenarios, is extremely difficult from a consent point of view. 
One could rent out public spaces, hire actors who give consent, and recreate various real-world scenarios.
Aside from the extremely high cost of such a reenactment, the resulting data would not be as natural as collected from a real public space.

A second solution is moving towards synthetic data, which is not only free of privacy issues, but also provides high-quality ground-truth annotations without the usual expensive and time consuming annotation effort.
Several works~\cite{DBLP:conf/eccv/RichterVRK16, DBLP:conf/corl/DosovitskiyRCLK17, fabbri2018learning, DBLP:conf/cvpr/KimWLK20, motsynth} use video game simulations to create detection, segmentation, and tracking training data.
Nonetheless, all existing synthetic datasets suffer to some degree from the synthetic-to-real domain gap caused by poor simulation quality, unrealistic behavior or differences in the underlying image-data distribution. 
Hence, to bridge the domain gap many methods still benefit from fine-tuning on real-world input data.

A third solution is to record the data without consent, but only release it after it has been de-identified.
Obfuscation or blurring is not an option for the generation of machine learning datasets cause, as we show in our experiments, training models for pedestrian downstream tasks on such data deteriorates the inference performance on not de-identified test data.
The authors of~\cite{yang2021imagenetfaces} created a version of the popular image classification dataset ImageNet~\cite{ILSVRC15} with obfuscated, thereby de-identified, faces and measured only a small drop in classification accuracy.
More recent approaches~\cite{DBLP:conf/iccv/Gafni19, DBLP:conf/cvpr/MaximovEL20} improve the performance of de-identified training data by synthesizing faces with Generative Adversarial Networks (GANs).
While promising, existing methods do not resolve the data privacy issues in pedestrian datasets as they suffer from two main issues: (i) Insufficient privacy guarantees or identity leakage by training the model on unethically collected image data, and (ii) face de-identification alone still allows pedestrian identification by their gait or clothes~\cite{chang2006identification, oh2016faceless}.

In this work, we advocate for full-body de-identification as the proper way to achieve data privacy for human-centered data collection, and focus on the case of pedestrian training datasets for detection, tracking, and re-identification.
To this end, we introduce the \emph{Pedestrian dataset de-identification} (PDI) task and motivate the necessity for full-body de-identification.
The task introduction includes relevant downstream tasks and their evaluation metrics.
We hope that the PDI task inspires future research on dataset de-identification in a competitive manner.
We propose \methodname, a full-body de-identification pipeline that serves as a first baseline for addressing PDI.
In order to prevent identity leakage during de-identification, we propose a two-stage de-identification method: (i) a Generative Adversarial Network (GAN) that transfers synthetic identities to real images, synthesizing pose- and temporal-consistent de-identified identities, and (ii) a second GAN that adapts the synthesized identities to real scenes, significantly improving their applicability as training data for downstream task models.
Most importantly, \methodname is never trained on any real identities.

To demonstrate its effectiveness, we provide \emph{\methodname17} a de-identified version of the MOT17~\cite{MOT16} multi-object tracking dataset.
\methodname17 is not only \textit{free from data privacy issues} but as shown by our experiments provides adequate MOT training data.
To this end, we train a variety of modern recognition methods and achieve competitive results in pedestrian detection, re-identification, and tracking.
We believe full-body de-identification will narrow the gap between synthetic and real data in a \emph{privacy-conscious} manner.
Furthermore, we hope this work fosters the awareness of the computer vision community and inspires future research into ethical data collection and de-identification.

In summary, our \textbf{contributions} towards more data privacy in pedestrian datasets are: 
\begin{itemize}
    
    \item We draw attention to the privacy issues in the data collection of 
    large-scale human and pedestrian datasets.

    \item We motivate the necessity for full-body de-identification and introduce the Pedestrian Dataset De-Identification (PDI) task, which tackles the generation of de-identified training data.

    \item We propose \methodname, a two-stage full-body de-identification pipeline which avoids identity leakage by replacing original image data with synthetic identities and improves downstream task performance by adapting the synthetic data to the original image domain.
    
    \item We demonstrate the effectiveness of \methodname by providing a de-identified version of MOT17. 
    The \mbox{\methodname17} dataset is the first MOT dataset free from privacy issues which achieves competitive results in downstream tasks such as detection, re-identification, and tracking.

\end{itemize}

\section{Pedestrian dataset de-identification }
\label{sec:datasetanonymization}

We present the \emph{Pedestrian Dataset De-identification} (PDI) task which aims at the creation of de-identified pedestrian datasets void of any data privacy issues and
applicable for training machine/deep learning models on a range of downstream tasks.
In addition to a potentially problematic data collection process, we recognize that pedestrian datasets facilitate the development of controversial surveillance applications.
Nonetheless, detecting and tracking pedestrians is crucial for safe autonomous vehicles, hence, we believe this research has an important benefit for society. A further discussion on this topic is out of scope of the paper.
For most vision tasks, progress is achieved by training, testing and comparing methods on fixed datasets/benchmarks.
The nature of the PDI task differs from the common evaluation pipeline as it includes the creation of the training data itself.
Hence, we not only compare the task performance obtained with said data but also the quality, in our case degree of de-identification, of the data.
Following, we will discuss several aspects of the multi-metric PDI task, motivate full-body de-identification and introduce a range of downstream pedestrian tasks.

\noindent \textbf{Privacy guarantees.}
To avoid any privacy issues, neither the training nor inference of a dataset de-identification pipeline should get in touch with any unethically or unlawfully collected data. 
This includes, in particular, any identifiable data related to the target identities which are supposed to be de-identified.
Existing de-identification methods~\cite{DBLP:conf/iccv/Gafni19, DBLP:conf/cvpr/MaximovEL20} fail to provide those privacy guarantees by training their models on person data collected without explicit consent.
Furthermore, their inference pipeline does not guarantee that there is no leaking of person identities into the generated de-identifications.
The lack of ethically collected data complicates a data privacy-conscious development of pedestrian de-identification methods.
A possible solution comes in the form of synthetic data which is by design free from privacy concerns.
Although not yet explored for dataset de-identification, existing synthetic datasets~\cite{DBLP:conf/eccv/RichterVRK16, DBLP:conf/corl/DosovitskiyRCLK17, fabbri2018learning, DBLP:conf/cvpr/KimWLK20, motsynth} have shown great potential as direct replacements for real pedestrian data.
%
%
%
%
%
An additional benefit of synthetic data is its ability to avoid biases, \eg, with respect to gender or race, existing in most real-world datasets.

\noindent \textbf{Why full-body de-identification?}
Modern de-identification methods~\cite{DBLP:conf/iccv/Gafni19,DBLP:conf/cvpr/MaximovEL20} and data privacy regulations, such as the GDPR~\cite{EUdataregulations2018}, are mostly concerned with face de-identification.
However, in particular for pedestrian data, we argue that face de-identification alone is not able to provide sufficient privacy guarantees.
A key characteristic of pedestrian data is its documentation of humans and their natural movement/behaviour in public spaces.
This not only makes it more difficult to ask for consent during the data collection process but makes the data particularly vulnerable for identification.
Recording humans in a private context facilitates the identification beyond facial features.
For example, current research~\cite{chang2006identification, oh2016faceless} shows that it is possible to infer a person's identity using body features such as dress, gait, and body shape.
Modern de-identification methods should strive to remove as many identifiable features as possible.
As a first step, this could be recognized by focusing research from face to full-body de-identification methods.
Furthermore, historically the identification via facial features~\cite{eigenfaces1991, guillaumin2009you, DBLP:conf/cvpr/SchroffKP15, deepface2016} has been well-studied before legal regulations, such as GDPR, took action on it.
By advancing the full-body de-identification technology, we hope that regulations catch up and recognize its importance.

\noindent \textbf{Training data generation.}
Existing identity de-identification methods are usually invasive and lead to some form of image degradation, \eg, from blurring. 
For a wide range of applications, \eg, for the creation of street views in digital maps, these alterations do not pose any problems.
However, for the generation of de-identified training data unrealistic image regions, visible artifacts or even changes to the underlying image distribution can severely deteriorate the performance of downstream tasks.
This was for example discussed in~\cite{yang2021imagenetfaces} and we further analyze the effects of different de-identification techniques in~\cref{subsec:baselines}.
All findings agree that existing full-body de-identification methods, \eg, via blurring, negatively impact the performance of deep learning models trained on this data.
Full-body pedestrian dataset de-identification in real-world scenarios faces several challenges.
The identity replacements must consider temporal consistency, object occlusions and view point changes.
Furthermore, the de-identification of an identity requires its recognition within the data via ground truth labels or model predictions.
Obtaining this information is particularly difficult and/or expensive for pedestrian data.

\noindent \textbf{Pedestrian downstream tasks.}
The computer vision community works on a range of challenging human-centered recognition tasks.
For the scope of our newly introduced PDI task, we want to focus on pedestrian recognition which faces the most severe data privacy issues.
In particular, we want to focus on multi-object tracking data as it incorporates several downstream tasks which demand different properties of a sufficient de-identification.
We will introduce the PDI task on the example of MOT17~\cite{MOT16}.
A de-identified and data privacy-conscious version of~\cite{MOT16} must provide adequate training data for the detection, re-identification, and tracking pedestrian tasks.

\subsection{Tasks \& Metrics}
The PDI task is composed of de-identification task and several downstream image and video tasks.
All of which allow us to measure different aspects of the de-identification with their individual sets of evaluation metrics.
While the de-identification quality of the generated data can be measured directly, each downstream task requires a separate training of task-specific models.
To ensure the validity of any results, we recommend to train a range of different methods for each downstream task.
In the following, we will introduce each PDI subtask, explain its challenges with respect to a successful de-identification and introduce relevant metrics.
For more technical details and a list of metrics mentioned in this work, we refer to the supplementary.

\noindent \textbf{De- and re-identification.}
A successful de-identification method must reliably replace all original identities from the target data and be robust against reverse engineering attempts.
Re-identification instead measures how well we can identify a person throughout its trajectory over a sequence.
Both identification tasks are evaluated with query-gallery experiments which test if a query identity can be successfully matched to a set of diverse gallery pedestrians.
The query and gallery samples consist of bounding box crops around each identity.
The task performance is measured via the Cumulative Matching Characteristics (CMC) rank-$k$ which averages the number of positively matched queries in the top-$k$ gallery samples.
For the de-identification quality, we perform a modified re-identification, \ie, de-identification query-gallery experiment.
To this end, we de-identify the query set and penalize the identification with any of their corresponding original gallery samples.
A successful deID experiment shows that it was not possible to recover the original identity from the de-identified dataset.
The downstream re-identification task poses significant challenges for any PDI method as it relies on temporal consistent de-identifications which are stable throughout the sequence and sufficiently distinct for each target identity.

%


%

\noindent \textbf{Object detection.}
The detection of pedestrian objects in the given image or video data is at the core of most pedestrian-related applications.
An object detector must predict bounding box coordinates encompassing each object and work reliably for varying object size and appearance as well as from different view points.
We evaluate a trained detector by computing the Average Precision (AP) across test data.
This metric measures how many of the detected objects are relevant, \ie, focuses on the avoidance of false positives.
For de-identified detection training data, we expect more realistic de-identifications to benefit the downstream task generalization, \eg, by maintaining human body proportions.
Since we evaluate single images/frames, temporal consistency is less important than the properties of the object-background or object-object boundaries.
The generation of the latter is particularly challenging for any de-identification method.

\noindent \textbf{Multi-object tracking.}
The multi-object tracking (MOT) downstream task consists in itself of two subtasks: (i) The detection of pedestrians in each frame and (ii) an association of identities to tracks across a sequence of frames.
Popular tracking-by-detection approaches decouple those steps by applying separate detecting and association models/steps.
A MOT method is evaluated by reporting the Multi-Object Tracking Accuracy (MOTA) and Identity F1 (IDF1) score.
The former metric focuses on object coverage, \ie, detection performance, by combining false positives, missed targets, and identity switches.
The IDF1 score measures identity preservation via the ratio of correctly predicted identity tracks over the average number of ground truth and computed detections.
MOT has the same requirements as detection and reID with respect to its (de-identified) training data.
These requirements are only exacerbated for frame-to-frame de-identifications of objects with strong occlusions and non-linear movement.

\section{Related work}
\label{sec:related_work}

\subsection{Datasets and privacy} 
Most popular person detection~\cite{COCO, crowdhuman}, re-identification \cite{ristani2016MTMC, zheng2015scalable, li2014deepreid}, and tracking~\cite{MOT16, DBLP:conf/cvpr/GeigerLU12}  datasets consist of recorded and annotated real scenes with real humans.
The creation of such datasets should require consent from every identifiable person in the data. 
Over the last years, the awareness of data privacy has steadily increased in the computer vision community.
For example, the authors of~\cite{asano21pass} released a version of ImageNet excluding people. 
In another notable case, due to consent mishandling, the Duke MTMC dataset \cite{ristani2016MTMC}, one of the most prominent person re-id datasets, was retracted.
Furthermore, recent editions of top machine learning (NeurIPS 2021) and computer vision (CVPR 2022) conferences require to discuss the ethical aspects of new datasets along with their creation and licensing process. 
The aforementioned problems can be circumvented through the use of synthetic data for training~\cite{motsynth, fabbri2018learning, varol17_surreal, wood2021fake}.
These datasets are not only free from any data privacy concerns but also provide perfect ground truth without the annotation effort.
However, despite the constant progress in the computer graphics community, improving the visual and behavioral realism of human and scene modeling is in itself a challenging and labor-intensive endeavor. 
When training exclusively on synthetic datasets, there is still a synthetic-to-real domain gap, resulting in sub-optimal test-time performance.
An example is the MOTSynth~\cite{motsynth} dataset which provides promising results for MOT downstream tasks but requires fine-tuning on real tracking datasets, such as MOT17 \cite{MOT16}, to achieve competitive results.
We propose to leverage synthetic data in a different way: We transfer synthetic identities to real data in order to achieve de-identification. 
By doing so, we provide a bridge between synthetic and real-world data, ensuring good de-identification and competitive results when training models without using any data with real identities.

\begin{figure*}[ht]
\begin{center}
\includegraphics[width=0.9\textwidth, trim={0px 200px 80px 0px}, clip]{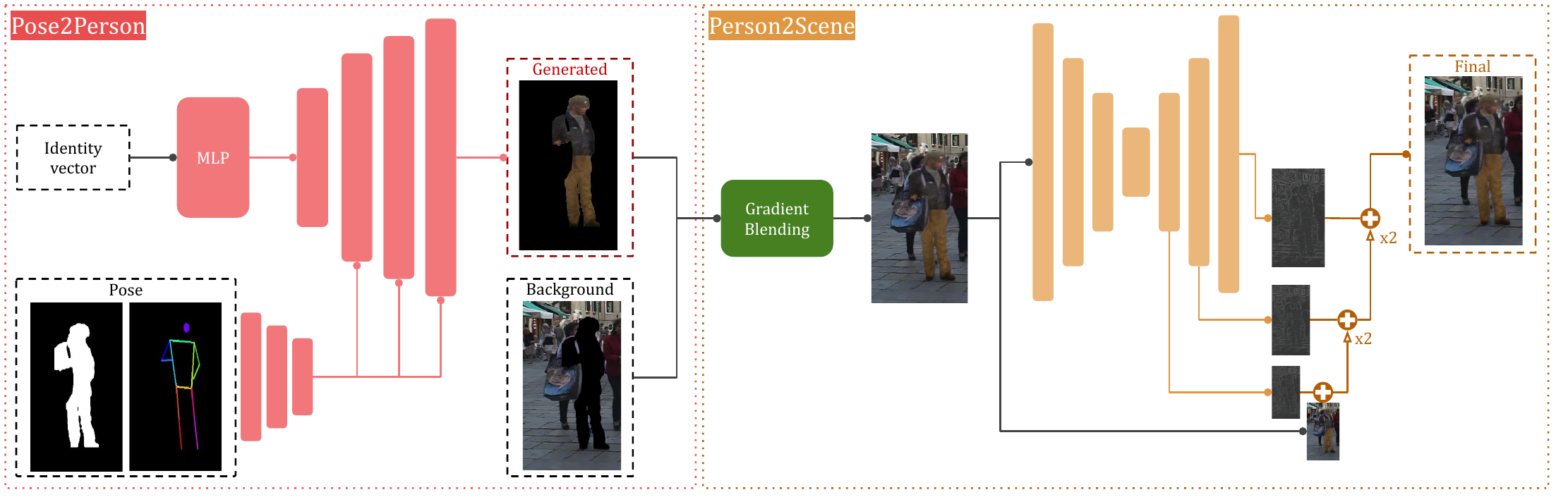}
\end{center}
\vspace{-.5cm}
   \caption{We present \methodname, a two-stage de-identification framework for pedestrian recognition datasets.
   The shown inference pipeline first generates an anonymous identity by providing the original pose to the \netone.
   In the second stage, we apply gradient blending and \nettwo to adapt the generated identity into the real image domain.
   }
\label{fig:pipeline}
\end{figure*}

\subsection{Identity de-identification} 
An additional measure to deal with data privacy issues is the de-identification of real data.
Existing de-identification methods are largely focused on obfuscating facial features \cite{DBLP:conf/cvpr/GrossSTB06, DBLP:conf/cvpr/SunMOGSF18, DBLP:conf/eccv/SunTXFTS18, DBLP:journals/corr/abs-1909-04538, ren2018privacy, DBLP:conf/iccv/Gafni19, DBLP:conf/cvpr/MaximovEL20, DBLP:conf/cvpr/ChenCYL21, DBLP:conf/iccv/Cao21, DBLP:conf/accv/MaximovACCV22}.
While these methods achieve high de-identification rates for faces, they still do not resolve the important data privacy concerns for pedestrian recognition datasets.
For example, identifiable body features such as clothing or tattoos in conjunction with scene information can be sufficient to identify pedestrians on the scene.
Furthermore, most aforementioned methods do not guarantee full de-identification since they still use the real image and real identity to generate the de-identified version. It is therefore not possible to quantitatively measure identity leakage. 
\cite{DBLP:conf/eccv/SunTXFTS18} uses rendered faces as substitutes, but still trains on real identities to create the final realistically-looking face.
We aim at overcoming these  problems of de-identification methods by first proposing a de-identification model for full bodies, and further guaranteeing privacy by using synthetic identities as a replacement for real ones. 
Our only constraint to guarantee full de-identification is the access to high-quality annotations, such as segmentation masks, hence, our work is focused on de-identifying training datasets rather than as a solution for on-the-fly de-identification of test data.

\subsection{Body synthesis} 
Recent advances in image~\cite{DBLP:conf/cvpr/IsolaZZE17, DBLP:conf/iccv/ZhuPIE17, DBLP:conf/cvpr/Park0WZ19, DBLP:journals/corr/abs-2109-06166} and video synthesis~\cite{DBLP:conf/nips/Wang0ZYTKC18} use conditional generative adversarial networks (GANs) for the generation of human bodies and faces.
Body synthesis and re-targeting methods~\cite{DBLP:journals/corr/abs-2109-06166, men2020controllable} create realistic images in a variety of poses.
However, they work in controlled settings and require additional expensive input such as body part segmentation or even real source images.
Motion transfer works~\cite{chan2019everybody} produce high-quality temporally consistent videos but are usually restricted to fully visible persons with little occlusion. 
While these methods produce realistically looking synthesized humans, their application as training data for modern pedestrian recognition models remains to be explored.
The first step in this direction has been the augmentation of person re-identification datasets~\cite{zou2020joint, zheng2019joint}.
This is achieved by training a conditional GAN to swap the appearance of clothes between different identities within a given dataset.
This type of alteration of the real data does not satisfy our requirements on data privacy, since we still have access to the original identities in the dataset.
In a similar way, CIAGAN \cite{DBLP:conf/cvpr/MaximovEL20} gives \textit{preliminary} results in body de-identification, however the data generated by them is not useful for training, in addition to not satisfying the requirements on data privacy, \textit{e.g.,} using real identities.
In order to prevent this issue, we transfer the synthetic identities to the real ones, eliminating the possibility of reverse engineering.
We expect our work to demonstrate how existing public datasets can be substituted by completely de-identified versions which allow models to achieve comparable performance on real person detection, re-identification, and tracking data.

\section{\methodname}
\label{sec:method}

We tackle dataset de-identification by generating replacements for pedestrians and then seamlessly merging those replacements into the scene. 
Our full-body de-identification pipeline \emph{\methodname} illustrated in Fig.~\ref{fig:pipeline} consists of two stages.
In the first stage, \stageone, we replace pedestrians by generating a synthetic, anonymous counterpart conditioned on the original pedestrian body pose.
However, as we show in the ablation studies (see~\cref{subsec:ablation}), a network trained only on de-identified data from \stageone reaches significantly worse results on real data.
The poor results are caused by the uneven boundaries between background and the generated identities, and the domain gap between synthetic and real data. 
To target these issues, we employ the second stage of our pipeline, \stagetwo.
We smooth the transition between the background and the newly generated pedestrian using gradient blending and adapt the generated part to match the target domain distribution using \nettwo.
In order to learn to adapt to the target distribution, we train \nettwo on real data, however, we remove all pedestrians from the data.
Hence, our entire de-identification pipeline is trained in a privacy-protected way - without using any real identities and instead learning to substitute them with synthetic counterparts.
We now describe our model, its architecture, and training objectives.

\subsection{Pose2Person}
The first stage of our pipeline is an image translation neural network trained with GAN algorithm~\cite{DBLP:conf/cvpr/Park0WZ19}. 
Our objective for \netone (P2P) is to synthesize an image of a pedestrian given: (i) a mask and a pose of the original pedestrian, (ii) a chosen synthetic identity,  see~\cref{fig:pipeline}. 
We use the instance mask and the body joint positions as the first input to avoid using any appearance information of real individuals. 
The second input we need is an identity in order to define the specific appearance we want to generate. This is important to generate a consistent appearance across multiple frames, \ie, the entire pedestrian track.
We propose training the network fully on the synthetic dataset \cite{motsynth}. From there, we extract images of $M$ identities based on their bounding boxes along with their pose information. 
We use these identities as our input condition by encoding identity numbers as one-hot vectors. The network then learns to transfer these identities to new poses.
Finally, to generate pose-aware outputs, we condition the discriminator on the body joints. 

As a backbone, we use spatially-adaptive denormalization (SPADE) \cite{DBLP:conf/cvpr/Park0WZ19} layers as building blocks of the network.
The generator outputs a masked generated pedestrian which is then combined with the image background and passed to the conditional discriminator. 
We use LSGAN~\cite{DBLP:conf/iccv/MaoLXLWS17} loss function to train \netone and we couple it with additional reconstruction losses. The reconstruction losses help with generator convergence and introduce a consistent output for the input identity condition.
The generator's loss is defined as:
\begin{equation}
    \resizebox{0.9\columnwidth}{!}{
    $\underset{G}{min} V (A) = LSGAN(b, G(i))  + L1(I, G(i)) + VGG_P(I, G(i))$,
    }
\end{equation}
where $VGG_P$ represents the perceptual loss of VGG net \cite{DBLP:conf/cvpr/GatysEB16}, $I$ is the original image, $b$ is the real data label, $i$ is the input to the generator $G$, and $D$ is the discriminator.

\begin{figure}[t]
    \centering
    \includegraphics[width=\columnwidth, trim={0px 0px 115px 0px}, clip]{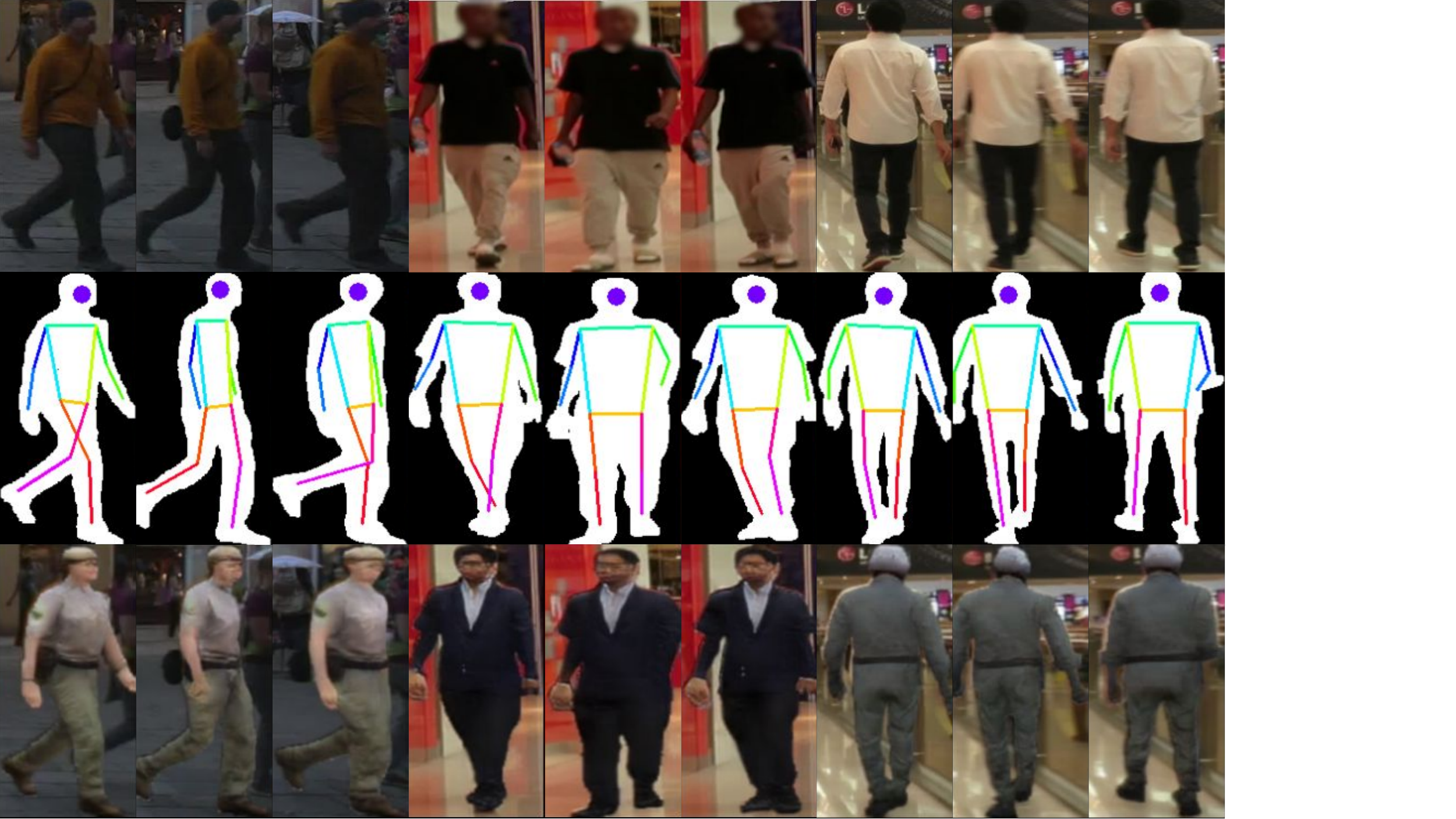}
    
    \vspace{-.3cm}
  \caption{\small Qualitative results of our de-identification method. In the first row we show the original images from different times frames, identities and sequences. 
   The second row shows their respective pose information used as an input to our pipeline. The third row shows our generated de-identification adapted to scenes.    
   }
    \label{fig:details}
    \vspace{-0.3cm}
\end{figure}

\subsection{Person2Scene}
The second stage of our pipeline concentrates on merging synthesized pedestrians into real scenes. It consists of an image gradient blending followed by our adaptation neural network.

\noindent \textbf{Gradient blending.} To combine the generated and original images, we can simply paste a generated person to the real background using its mask. However, such a naive technique leads to noticeable transition artifacts between image borders. 
If we use these images to train a deep learning model, it will quickly learn to exploit such artificial boundaries for its task, failing to generalize to natural scenes, as we show in the ablation studies.
We can prevent this issue by creating a more seamless transition between both images.
To achieve it, we use the Laplacian pyramid blending method \cite{pyramidblending} to mix gradients between the background and the generated person. We mix Laplacian pyramids of both images on several levels and then collapse them into a final mixed image. 
This blending results in more natural boundaries and better results as shown in the ablation~\ref{subsec:ablation}. 

\noindent \textbf{Adaptation network.} 
Since our \netone is trained on synthetic pedestrians, the generated output will be close to that synthetic domain. Blending helps us with the boundary regions, but there is still a domain gap between a generated person and a real background. 
To bridge that gap, a straightforward approach is to learn how to adapt from synthetic to real pedestrians. 
However, we have an important constraint: we cannot train on real identities. We instead propose to train only on the real background to learn to synthesize general images of the real distribution.

The \nettwo (P2S) adaptation network is inspired by Laplacian pyramids~\cite{pyramidblending} and trained for image reconstruction.
The network encodes the input image and decodes multiple scales of the Laplacian pyramid which we collapse into the reconstructed image.
Intuitively, the network reconstructs image details in the trained domain while preserving the overall appearance of the generated image.
Furthermore, we additionally apply augmentations to the input image, e.g., blurring and sharpening to force \nettwo to not simply reconstruct the image by relying on the input, but to hallucinate the missing details to recover the original image. 
We show in the ablation study that such an adaptation improves performance further. 

The \nettwo architecture is a U-Net~\cite{DBLP:conf/miccai/RonnebergerFB15} composed of residual blocks. 
Instead of directly reconstructing the input image, the decoder predicts its Laplacian Pyramid. 
Each residual block in the decoder outputs a Laplacian of one pyramid scale. 
To compose the final image, we use the last level of $n$-level Gaussian pyramid as our initial low-pass image representation and the predicted Laplacian Pyramid.
This design synthesizes image details and is intended to match the underlying image distribution.
The total loss of the network is defined as:
\begin{equation}
    \resizebox{0.9\columnwidth}{!}{%
    $\underset{A}{min} V (A) = LPIPS(j, A(j))  + L2(j, A(j)) + SSIM(j, A(j))$,
    }
\end{equation}
where $A$ represents the \nettwo adaption network, $j$ is the input image patch with masked out pedestrians, $LPIPS$ denotes a learned perceptual patch similarity loss~\cite{zhang2018perceptual}, and $SSIM$ -- structural similarity loss~\cite{wang2004ssim}. $LPIPS$ is similar to $VGG_P$, but attuned more to perceptual differences between images. Additionally, we added $SSIM$ loss to put focus on structural details which can be ignored by perceptual-based losses.

\section{Experiments}
\label{sec:experiments}

To demonstrate the effectiveness of \methodname as a first baseline for the Pedestrian Dataset De-Identification (PDI) task, we evaluate its de-identification performance as well as its ability to generate training data for multi-object tracking (MOT) tasks. 
To this end, we de-identify the \emph{MOT17}~\cite{MOT16} dataset and create its data privacy-conscious counterpart \textit{\methodname17}.
First of all, we ablate the design choices of our \methodname pipeline and then compare it with several de-identification baselines.
Finally, we analyse the synthetic-to-real (syn2real) gap when pretraining on MOTSynth~\cite{motsynth} and demonstrate detection and tracking benchmark results comparable with state-of-the-art on the MOT17Det~\cite{MOT16} and MOT17~\cite{MOT16} datasets without training on any identifiable pedestrian data.

\subsection{Evaluation details}
To demonstrate the robustness of our method across different experiments, we train and evaluate multiple models/methods per downstream task.
The MOT17~\cite{MOT16} dataset contains 7 training and 7 test sequences which depict different urban scenes and vary in length, number of labeled pedestrian objects and camera viewpoint.
Except from the benchmark evaluations, we report metrics averages over a $7$-fold cross-validation split on the MOT17 training sequences, \ie, a single split consists of 6 train and one test sequence.
It should be noted, that we are only interested in the ability of \methodname to generate training data.
Hence, we train on various types of input data (original, de-identified or synthetic MOTSynth~\cite{motsynth} sequences) but always test each split on the unaltered original sequences.
If not stated otherwise, the presented models are all pretrained on the ImageNet~\cite{ILSVRC15} dataset.
The arrows in our tables indicate low or high optimal metric values.
For more evaluation details of the individual PDI downstream tasks, a list of all metrics in the following tables and additional technical details of our \methodname pipeline, we refer to the supplementary.

\noindent \textbf{Creation of \methodname17.}
We train~\netone of our \methodname pipeline on a set of selected synthetic MOTSynth~\cite{motsynth} identities.
In order to de-identify MOT17~\cite{MOT16}, we map each original identity to one of these selected identities.%
To de-identify a single identity, the~\netone receives its mapped MOTSynth identity vector and frame-wise poses consisting of instance masks and keypoints (see~\cref{fig:pipeline}).
The entire \methodname pipeline is applied on a frame-by-frame basis, processing each frame independently.
Our \netone model requires high quality pose inputs, \ie, masks and joints.
However, the MOT17~\cite{MOT16} benchmark only provides ground truth bounding boxes.
The MOTS20~\cite{MOTS} dataset provides ground truth masks but only for a subset of the MOT17 sequences and objects.
For the missing objects and joints, we rely on predictions from state-of-the-art recognition models.
Training with spatial augmentations (cropping/resizing), allows the~\netone to generalize to occluded and/or predicted pose inputs.

\begin{table}[t]
\resizebox{1.0\columnwidth}{!}{%
\begin{tabular}[t]{l|ccc|cc|c}
\toprule
\multicolumn{1}{c|}{\multirow{2}{*}{\methodname}} & \multicolumn{3}{c|}{Input pose} & deID           & reID           & Faster-RCNN \\ \cmidrule{2-7} 
\multicolumn{1}{c|}{} & \rotatebox[origin=c]{90}{Mask}   & \rotatebox[origin=c]{90}{Joints}   & \rotatebox[origin=c]{90}{SMPL}   &
\multicolumn{2}{c|}{CMC rank-1 $\uparrow$} & AP $\uparrow$         \\ 
\midrule

\multirow{3}{*}{Pose2Person (P2P)}    & \checkmark & -- & -- & 84.1 & 63.7 & 41.3 \\
 & \checkmark & \checkmark & --  & 83.1 & 64.3 & 46.9 \\
 & \checkmark & \checkmark & \checkmark & 81.7 & 62.4 & 46.6 \\ 
 
\midrule
P2P + Blending (B) & \checkmark & \checkmark & -- & 82.3   & 65.3 & 52.0 \\
P2P + B + Person2Scene (P2S) & \checkmark & \checkmark & -- & 81.5 & 66.3 & 60.2 \\
P2P + B + P2S + Aug. & \checkmark & \checkmark & -- & 82.7 & 65.1 & 65.3 \\

\bottomrule
\end{tabular}
}
\vspace{-0.2cm}
\caption{
        Ablation study on the components of our {\methodname} de-identification pipeline.
        We measure their effect on de-identification (deID), reID and Faster R-CNN~\cite{rennips2015} detection performance.
        }
        \label{tab:ablation}
        \vspace{-0.4cm}
\end{table}

\subsection{Ablation study}
\label{subsec:ablation}

We ablate individual components of \methodname in~\cref{tab:ablation}.
For the baseline~\textit{Pose2Person} (P2P) stage, we only use masks as input pose and paste the generated anonymous identity directly into the frame.
This model already achieves satisfying de- and re-identification performance but fails to provide sufficient training data for detection.

\begin{table*}[ht]
    \begin{center}
        
    \resizebox{\textwidth}{!}{
    \begin{tabular}[t]{l | P{1cm}P{1cm} | P{2cm} P{2cm} | P{2.3cm} P{3.0cm} H P{2.3cm}}
        \toprule
        
        \multirow{4}{*}{\shortstack[c]{De-identification\\ method}} & deID & reID & \multicolumn{2}{c|}{Detection} & \multicolumn{4}{c}{Multi-object tracking}  \\
          \cmidrule{2-9}
         &  \multicolumn{2}{c|}{ResNet50} &  FRCNN~\cite{rennips2015} & RetinaNet~\cite{lin2017focal}  & Tracktor~\cite{tracktor} & Tracktor~\cite{tracktor} + reiD &
         MPNTrack &
         CenterTrack~\cite{center_track}\\
        \cmidrule{2-9}
        & \multicolumn{2}{c|}{CMC rank-1 $\uparrow$} & \multicolumn{2}{c|}{AP $\uparrow$} &  \multicolumn{4}{c}{MOTA / IDF1 $\uparrow$} \\

        \midrule

        None   & 0.0    & 68.9 & 69.3 & 71.1 & 30.4 / 36.0 & 31.2 / 40.0 & 62.3 / 64.7 & 37.5 / 41.1 \\
        Black-white blur      & 66.5 & 58.1   & 35.6 & 48.5 & -- & -- & XX & -- \\
        
        White cutout     & 85.3   & 49.3   & 9.4 & 29.4 & -- & -- & -- & -- \\
        
        CIAGAN~\cite{DBLP:conf/cvpr/MaximovEL20}    & 78.9 & 60.9 & 46.5 & 47.8 & -- & -- & -- & -- \\
                
        \textbf{\methodname}     & 82.7 & 65.1 & 65.3 & 65.6 & 25.4 / 35.1 & 25.6 / 35.4 & XX & 26.9 / 28.3 \\

        \bottomrule
    \end{tabular}    
    }
    \vspace{-0.3cm}
    \caption{\small 
        We demonstrate the effectiveness our proposed de-identification method with respect to several handcrafted baselines on a range of MOT related tasks, namely, re-identification (reID), detection and tracking.
        Furthermore, we measure the de-identification (deID).
    }
    \label{tab:baselines}
    \end{center}
    \vspace{-0.5cm}
\end{table*}

\noindent \textbf{Input pose.}
We compare different input poses for the P2P generation stage. 
To outline the synthesised area, all versions include the object segmentation mask. 
We increase detection performance by adding joints (P2P-J) to the input pose.
The decrease in deID performance is justified by the overall 10.7 points increase in detection.
However, 2D joints are ambiguous and not sufficient to fully determine a human pose including body part occlusions.
Hence, we experiment with an additional 3D model input. 
The SMPL~\cite{SMPL-X:2019} model can be fitted to a given set of 2D keypoints. 
While conceptually promising, the 3D model input gave no additional improvement for detection. 
In fact, the de- and reID performance got worse.
This is most likely due to noise and errors in the predicted keypoints exacerbated by the 3D model fitting.
We leave the exploration of additional inputs, \eg, a SMPL version more robust to unreliable keypoints, for future research.

\noindent \textbf{Adaptation.}
The Pose2Person model generates anonymous but synthetic identities.
The resulting domain gap to the real background data limits the achievable detection performance. 
We obtain more smooth object border transitions, we apply gradient blending (B) which results in an improvement of 5.1 detection points. 
Applying the~\textit{Person2Scene} (P2P) adaptation further mitigates the domain gap by fusing the generated synthetic identities into the real image distribution.
Without random training augmentations (Aug.) the \nettwo merely applies the image distribution existing in the sequence without the ability to synthesize and generalize to the unseen generated person parts.
The full adaption stage (B + P2S + Aug.) results in a total improvement of 18.4 AP detection points.

\subsection{Comparison with other de-identification baselines}
\label{subsec:baselines}

We compare with two handcrafted (white cutout and black-white blur) baselines as well as a full-body adaptation of the CIAGAN~\cite{DBLP:conf/cvpr/MaximovEL20} face de-identification method.
Each baseline and our \methodname method de-identify persons based on the same target object input poses. 
In~\cref{tab:baselines}, we show that our method outperforms all presented baselines.

\noindent \textbf{White cutout.}
As expected,~\textit{white cutout} represents the de-identification upper limit of 85.3\%.
A perfect de-identification is prevented by several remaining identification cues : (i) similar backgrounds, (ii) missing de-identifications from incomplete input masks, and (iii) matching via the mask silhouette itself.
Since our \netone relies on input masks and joints, (iii) is by design the only unresolvable identification cue.
However, as expected white cutouts do not provide valuable training data for any of the MOT tasks.
In particular, the region proposal network of the Faster R-CNN~\cite{rennips2015} suffers severely, achieving only 9.4\% AP. 
Due to the insufficient detection performance, we refrained from running additional tracking experiments.

\noindent \textbf{Blurring.}
A de-identification via ~\textit{black-white blur}, on the other hand, improves downstream task performance significantly, \eg, FRCNN detection and re-identification gain 26.2 and 8.8 points, respectively.
However, the improved performance is only obtained by revealing more of the original input data which causes a 18.8 point drop for deID.

\noindent \textbf{CIAGAN de-identification.}
The \emph{CIAGAN}~\cite{DBLP:conf/cvpr/MaximovEL20} face de-identification pipeline follows a similar identity transfer approach as the \stageone stage of \methodname.
To apply CIAGAN for full-body de-identification, we train their generation model on the same synthetic data as our \netone.
The resulting de-identification performance in~\cref{tab:baselines} is comparable to our approach.
However, without a synthetic to real adaptation stage, \ie, a pipeline specifically designed to generate \emph{training} data, their downstream reID and detection performance suffer severely.
In conclusion, all presented de-identification baselines are not able to produce sufficient de-identified MOT training data.

The data generated by~\textit{\methodname} achieves a degree of de-identification close to cutout (82.7 vs. 85.3) and performance close to the original data across all MOT tasks.
We assume that the reID gap of 3.8 points to the original data is due to the predicted and therefore temporal inconsistent and noisy \stageone input pose (masks and keypoints).
The reID and detection performance gaps correlate with the IDF1 and MOTA tracking performance. 
In particular, for Tracktor (+reID) which is evaluated with the FRCNN and ResNet-50 models trained for the respective detection and reID tasks.
For CenterTrack~\cite{center_track}, which requires center point prediction within the de-identified identity, the performance gap to the original data increases.
However, as a first of its kind, our \methodname pipeline is able to generate a privacy-conscious dataset with adequate downstream task performance.

\begin{table*}[t]
    \begin{center}
    \resizebox{\textwidth}{!}{
    \begin{tabular}[t]{ll | P{2.2cm} | P{2cm} P{2cm} P{2cm} | P{2.3cm} P{3.0cm} H P{2.3cm}}
        \toprule
        
        \multicolumn{2}{c|}{\multirow{2}{*}{Training data}} & reID & \multicolumn{3}{c|}{Detection} & \multicolumn{4}{c}{Multi-object tracking}  \\
        \cmidrule{3-10}
        \multicolumn{2}{c|}{} &  ResNet50 &  FRCNN~\cite{rennips2015} & RetinaNet~\cite{lin2017focal}  & CenterNet~\cite{center_track} &
        Tracktor~\cite{tracktor} &
        Tracktor~\cite{tracktor} + reID &
         MPNTrack &
         CenterTrack~\cite{center_track}\\
        \cmidrule{1-10}
        \multicolumn{1}{c}{PT} & \multicolumn{1}{c|}{FT} & CMC rank-1 $\uparrow$ & \multicolumn{3}{c|}{AP $\uparrow$} & \multicolumn{4}{c}{MOTA / IDF1 $\uparrow$} \\

        \midrule
         
        COCO  &         \multicolumn{1}{c|}{--}                & -- & 80.3 & 78.1 & 74.7 & 38.2 / 44.0 & -- & XX & -- \\
        MOTSynth  &       \multicolumn{1}{c|}{--}                  & 58.3 & 79.2 & 77.3 & 68.3 & 42.9 / 43.8 & 43.3 / 45.3 & XX & 41.2 / 43.9 \\
        
        \midrule
        
        \multicolumn{1}{c}{--} & MOT17                            & 68.9 & 69.3 & 71.1 & -- & 30.4 / 36.0 & 31.2 / 40.0 & XX & 37.5 / 41.1 \\
        \multicolumn{1}{c}{--} & \textbf{\methodname17}              & 65.1 & 65.3 & 65.6 & -- & 25.4 / 35.1 & 25.6 / 35.4 & XX & 26.9 / 28.3 \\
        
        \midrule
        
        COCO  &       MOT17             & 68.9 & 85.0 & 81.7 & 77.5 & 47.0 / 44.4 & 48.2 / 51.1 & XX & --  \\
        MOTSynth & MOT17                    & 70.9 & 83.5 & 79.1 & 77.5 & 46.3 / 48.8 & 46.3 / 49.0 & XX & 48.4 / 50.2 \\
        MOTSynth & \textbf{\methodname17}      & 68.5 & 83.2 & 79.1 & 74.8 & 45.8 / 46.0 & 46.4 / 48.5 & XX & 45.5 / 47.8 \\

        \bottomrule
    \end{tabular}
    }
    \end{center}
    \vspace{-.5cm}
    \caption{\small 
        We demonstrate how \methodname helps to bridge the gap between synthetic and real training data.
        To this end, we evaluate models for re-identification (reID), detection and tracking and pre-train (PT) and/or fine-tune (FT) on real (COCO~\cite{COCO}, MOT17~\cite{MOT16}), synthetic (MOTSynth~\cite{motsynth}) and de-identified (\methodname17) data.
    }
    \label{tab:syn2real}
\end{table*}

\subsection{Synthetic-to-real without data privacy issues}

Over the last years, synthetic training data has shown great potential for MOT and its sub-tasks.
However, even for large datasets, such as MOTSynth~\cite{motsynth}, there is still a generalization gap to real data.
To reach top performance, models require an additional fine-tuning on the real MOT17~\cite{MOT16} dataset.
Hence, the problem of data privacy and consent persists.
In~\cref{tab:syn2real}, we fine-tune on our de-identified \methodname17 dataset and demonstrate its ability to narrow the synthetic-to-real gap in a data privacy-conscious manner.

For the reID task, fine-tuning on the original MOT17 dataset improves performance by 12.6 CMC points over the baseline MOTSynth model.
The \methodname method is able to narrow the gap by achieving 68.5 in CMC which corresponds to a 10.2 point improvement over MOTSynth alone but without training the model on identifiable pedestrian identities.
We see a similar pattern in object detection. 
In the case of Faster R-CNN~\cite{rennips2015}, our de-identified data makes the performance drop by only 0.3 points.
However, RetinaNet~\cite{lin2017focal} improves from both MOT17 and \methodname17 by 0.8 points, indicating that for detection,
MOT17 can be \emph{fully replaced by our de-identified dataset.}
Finally, we compare the results of three multi-object tracking methods.
The presented Tracktor versions both achieve performance competitive to fine-tuning on the original MOT17 and gain several points over the MOTSynth baselines.
For example, Tracktor without reID improves results by 2.9 and 2.2 points for MOTA and IDF1, respectively.
The addition of the reID model is beneficial for every training data combination.
In fact, our performance of Tracktor + reID is marginally better than the one fine-tuned on the original dataset.
Furthermore, we compare CenterTrack~\cite{center_track} results and while fine-tuning on \methodname17 leads to worse results than on MOT17, we still show a significant improvement over the model trained only on MOTSynth. 
We gain 4.3 points in MOTA, and 3.9 points in IDF1.

In conclusion, for all three tasks and five different methods, fine-tuning on our \methodname17 dataset significantly improves the results over MOTSynth alone.
While \methodname17 is not sufficient for training from scratch (see~\cref{tab:baselines}), it shines when the general task can be pretrained, \eg, on MOTSynth.
With the results being close to models trained on real data, 
\methodname marks an important step towards privacy-conscious training of CNNs.

\begin{figure}[t]
    \centering
    \captionsetup[subfigure]{labelformat=empty}
    \subfloat[MOT17-03]{
        \includegraphics[width=0.45\columnwidth]{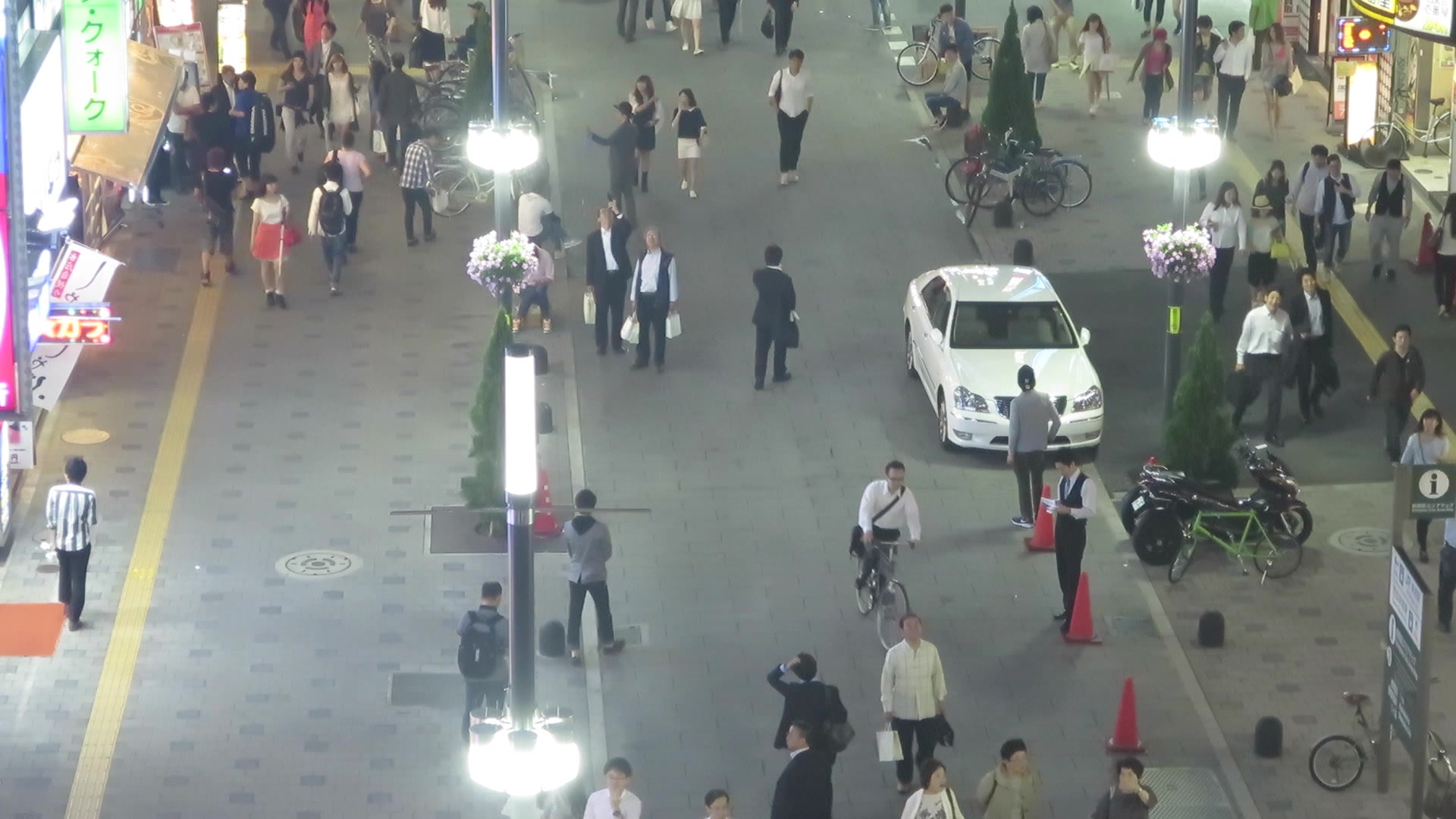}}
    \subfloat[MOT17-04]{
        \includegraphics[width=0.45\columnwidth]{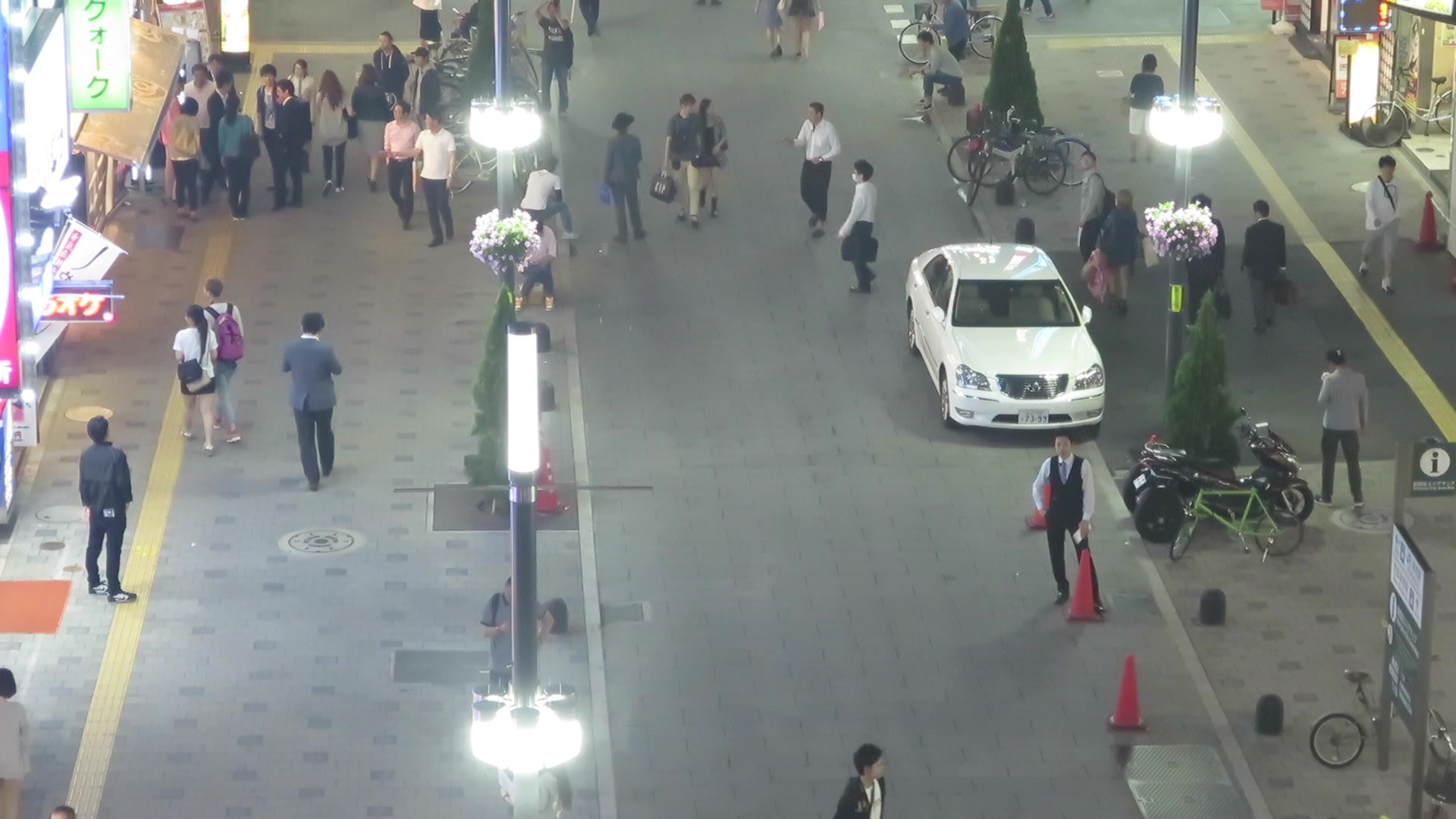}}
        \vspace{-0.1cm}
    \caption{
        \small The MOT17 train and test dataset~\cite{MOT16} each contains a sequence (\mbox{MOT17-03}, \mbox{MOT17-04}) picturing the same scene and several shared identities.
        Therefore, training models on the original and not de-identified data boosts their generalization to test.
        }
    \label{fig:mot17-03_vs_mot17-04}
\end{figure}

\begin{table}[t]
\begin{center}
\resizebox{\columnwidth}{!}{

    \begin{tabular}{l| ll| c c H |r r r H H}
    \toprule
    \multicolumn{1}{c|}{Method} & \multicolumn{1}{c}{PT} & \multicolumn{1}{c|}{FT} & AP $\uparrow$ & MODA $\uparrow$ & FAR $\downarrow$ & TP $\uparrow$ & FP $\downarrow$ & FN $\downarrow$ & Rcll. $\uparrow$ & Prcn. $\uparrow$  \\ [0.5ex] 
    \midrule

    DPM~\cite{dpmpami2009} & \multicolumn{1}{c}{?} & MOT17 & 61.0 & 31.2 & 7.1 & 78007 & 42308 & 36557 & 68.1 & 64.8\\
    FRCNN~\cite{rennips2015} & \multicolumn{1}{c}{?} & MOT17 & 72.0 & 68.5 & 1.7 & 88601 & 10081 & 25963 & 77.3 & 89.8\\
    SDP~\cite{SDP} & \multicolumn{1}{c}{?}  & MOT17 & 81.0 & \textbf{76.9} & \textbf{1.3} & 95699 & \textbf{7599} & 18865 & 83.5 & \textbf{92.6}\\
    ZIZOM~\cite{Lin2018GraininessAwareDF} & \multicolumn{1}{c}{--} & MOT17 & 81.0 & 72.0 & 2.2 & 95414 & 12990 & 19139 & 83.3 & 88.0\\
    YTLAB~\cite{cai16mscnn} & \multicolumn{1}{c}{?} & MOT17 & \textbf{89.0} & 76.7 & 2.8 & \textbf{104555} & 16685 & \textbf{10009} & \textbf{91.3} & 86.2\\
    
    
    \midrule
    
    FRCNN & MOTSynth & \multicolumn{1}{c|}{--} & 80.0 & 56.1 & 6.0 & 100015 & 35753 & 14549  & 73.7 & 87.3 \\
    
    FRCNN  & MOTSynth & MOT17 & 90.0 & 69.2 & 4.4 & 105019 & 25752 & 9545 & 91.7 & 80.3 \\

    FRCNN  & MOTSynth & \textbf{\methodname17} & 81.0 & 68.1 & 3.4 & 98343 & 20288 & 16221 & 85.8 & 82.9 \\
        
    \midrule
    \multicolumn{1}{c}{} & & \multicolumn{3}{c}{ FT without MOT17-04} \\ 
    \midrule
    
    FRCNN & MOTSynth & MOT17  & 81.0 & 59.1 & 5.9 & 102595 & 34931 & 11969 & 89.6 & 74.6 \\
    
    FRCNN & MOTSynth & \textbf{\methodname17} & 80.0 & 58.3 & 5.4 & 98690 & 31933 & 15874 & 86.1 & 75.6 \\

    \bottomrule
    \end{tabular}
    
}
\end{center}
\vspace{-0.5cm}

\caption{\small 
    We pretrain (PT) FRCNN on MOTSynth~\cite{motsynth} and evaluate different finetuning (FT) configurations on the MOT17Det~\cite{MOT16} detection test set.
    Removing sequence MOT17-04 from the finetuning training data closes the gap between our de-identification and the real data.
    A ? denotes unknown pretraining datasets.
    } \label{tab:mot17det_test}
    \vspace{-0.5cm}

\end{table}

\subsection{Benchmark evaluation}
\label{subsec:benchmark}
Finally, we demonstrate top test set results achievable by training on \methodname17 for both the MOT17Det~\cite{MOT16} detection and MOT17~\cite{MOT16} multi-object tracking benchmarks.
We limit our test set submission for~\cref{tab:mot17det_test} to the Faster R-CNN~\cite{rennips2015} (FRCNN) detector and train our own model on the MOTSynth dataset.
While synthetic data alone already achieves 79.0 AP, our fine-tuning on the MOT17 dataset boosts results to 90 AP.
If we apply our de-identified version, the AP improves by additional 2 points.
In general, we consider any improvement as a success since we simultaneously tackle the issues of data privacy.
However, these results stand in contrast to~\cref{tab:syn2real}, where we are on par with fine-tuning on the original data.
Analyzing the train and test sets, we see that the MOT17-04 train scene corresponds to the MOT17-03 test scene, see~\cref{fig:mot17-03_vs_mot17-04}, which is recorded a few minutes later.
The sequences even share several identities.
This overlap between train and test puts models trained on \methodname17 at a disadvantage as they can not capitalize on their similarities.
To understand the impact of those similarities, we repeat both fine-tuning experiments but remove the MOT17-04 sequence from the train set.
As expected, the overall performance drops, but the gap between fine-tuning on real or de-identified data shrinks significantly.

\begin{table}[t]
    \begin{center}
    \resizebox{\columnwidth}{!}{
    \begin{tabular}[t]{l| ll| cc HH rrr}
        \toprule
        \multicolumn{1}{c|}{Method}  & \multicolumn{1}{c}{PT} & \multicolumn{1}{c|}{FT} & MOTA $\uparrow$ & IDF1 $\uparrow$ & MT $\uparrow$ & ML $\downarrow$ & FP $\downarrow$ & FN $\downarrow$ & ID Sw. $\downarrow$ \\

        \cmidrule{1-10}

        TubeTK~\cite{tube_tk}                     &   JTA & MOT17   & 63.0 & 58.6 & 735  & 468 & 27060 & 177483 & 4137 \\
        CTracker~\cite{chained_tracker}     &    \multicolumn{1}{c}{--}     & MOT17      & 66.6 & 57.4 & 759  & 570 & 22284 & 160491 & 5529 \\
        CenterTrack~\cite{center_track}           &   CH   & MOT17 & 67.8 & 64.7 & 816  & 579 & \textbf{18498} & 160332 & 3039  \\

        QuasiDense~\cite{qdtrack}           & \multicolumn{1}{c}{--}  & MOT17  & 68.7 & 66.3 & 957  & 516 & 26589 & 146643 & 3378  \\

        TraDeS~\cite{Wu2021TraDeS}           &   CH  & MOT17  & 69.1 & 63.9 & 858  & 507 & 20892 & 150060 & 3555  \\

        GSDT~\cite{Wang2021126911}           &   6M   & MOT17 & 73.2 & 66.5 & 981  & 411 & 26397 & 120666 & 3891  \\

        PermaTrack~\cite{tokmakov2021learning}           &   CH+PD & MOT17   & 73.8 & 68.9 & 1032  & 405 & 28998 & 115104 & 3699  \\

        GRTU~\cite{Wang2021ICCVGRTU}           &   CH+6M  & MOT17  & 75.5 & \textbf{76.9} & 1158  & 495 & 27813 & 108690 & \textbf{1572}  \\

        TLR~\cite{wang2021multiple}           &   CH+6M & MOT17   & \textbf{76.5} & 73.6 & 1122  & 300 & 29808 & \textbf{99510} & 3369  \\
        
        \cmidrule{1-10}
        
        Tracktor + reID                      &   MOTSynth  &    \multicolumn{1}{c|}{--}    & 52.6  & 44.4 & 537 & 594 & 37692 & 219699 & 10320  \\
        Tracktor + reID                      &   MOTSynth  & MOT17  & 59.7 & 53.6 & 639 & 525 & 13266 & 206640 & 7512  \\
        Tracktor + reID                      &   MOTSynth  & \textbf{\methodname17}  & 55.3 & 47.9 & 549 & 564 & 13488 & 230445 & 8220  \\
        
        \cmidrule{1-10}

        CenterTrack                      &   MOTSynth  & \multicolumn{1}{c|}{--}  & 53.0 & 58.2 & 621 & 660 & 94458 & 164691 & 3517  \\
        CenterTrack                      &   MOTSynth  & MOT17  & 66.1 & 64.2 & 702 & 708 & 11856 & 176991 & 2292  \\
        CenterTrack                      &   MOTSynth  & \textbf{\methodname17}  & 57.1 & 59.4 & 576 & 834 & 28941 & 211050 & 2241  \\
        \bottomrule
        
    \end{tabular}
    }
    \end{center}
    \vspace{-.5cm}
    \caption{\small 
        Comparison of modern online multi-object tracking methods evaluated on the MOT17~\cite{MOT16} test set for private detections.
        We show that models pretrained (PT) on MOTSynth and finetuned (FT) on our \methodname dataset consistently outperform the synthetic pretraining alone.
        The pretraining datasets are abbreviated as: CH=CrowdHuman~\cite{crowdhuman}, PD=Parallel Domain~\cite{tokmakov2021learning} (synthetic), 6M=6 tracking datasets as in~\cite{zhang2021fairmot} and JTA~\cite{fabbri2018learning} (synthetic).
    }
    \label{tab:mot17_test}
    \vspace{-0.3cm}
\end{table}

\noindent \textbf{MOT17.}
In contrast to~\cite{motsynth}, we evaluate with private and not public detections as these are obtained by training on the original data.
This allows us to measure top achievable tracking performance when training on synthetic and/or de-identified data only.
For~\cref{tab:mot17_test}, we submit results for Tracktor + reID (which corresponds to Tracktor++~\cite{tracktor}) and CenterTrack, verifying their results from~\cref{tab:syn2real}.
Training Tracktor and the reID model on MOTSynth + \methodname17 outperforms the base MOTSynth model by 2.7 MOTA and 3.5 IDF1.
The same trend is observable for CenterTrack~\cite{center_track} which we improve by 4.1 MOTA.
It should be noted, that tracking-by-detection models such as Tracktor and CenterTrack also benefit from the similarities between train and test sequences observed for the MOT17Det detection experiments. 
Hence, there is still a gap to fine-tuning on the original MOT17.

\section{Conclusion}
We presented the full-body Pedestrian Dataset De-Identification task and a two-stage de-identification method \methodname.
Our method provides sufficient de-identification guarantees as it only relies on the mask and keypoint information of the original pedestrians.
To achieve this, we propose to replace identities with synthetic ones, generating pose- and temporal-consistent de-identifications.
Most importantly, our model is never trained on any unethically collected pedestrian identities.
Using our method we created \methodname17, a de-identified version of the MOT17 real-world multi-object tracking dataset. 
We showed training on our data allowed models to achieve competitive downstream performance for several MOT subtasks such as detection, re-identification, and tracking.
We hope this paper inspires future research and raises awareness for machine learning applications free of data privacy issues.
%

{\small
\bibliographystyle{ieee_fullname}
\bibliography{egbib_fat}
}

\end{document}